\documentclass[twocolumn,10pt]{article}
\usepackage{geometry}
\usepackage{graphicx}

\usepackage{siunitx}

\usepackage{hyperref}   
\hypersetup{
	colorlinks=true,
	linkcolor=blue,
	citecolor=blue,
	urlcolor=blue,
}
\usepackage{mathtools}
\usepackage{amsmath}
\usepackage{lipsum}
\usepackage{diagbox}
\usepackage{tabularx}
\usepackage{array}
\usepackage{epsfig} 
\usepackage{lipsum}
\usepackage{xspace}
\usepackage{amsmath}
\usepackage{booktabs}
\usepackage{caption}
\usepackage{xcolor}

\usepackage{algorithm}
\usepackage{algpseudocode}

\usepackage[style=ieee,backend=biber,hyperref=true]{biblatex}
\usepackage{authblk}

\usepackage{mathptmx}

\let\oldparagraph\paragraph

\renewcommand{\paragraph}[1]{\oldparagraph{#1.}}

\title{Deep Generative Design for Mass Production}
\date{}

\author[1]{Jihoon Kim$^*$}
\author[2,3]{Yongmin Kwon$^*$}
\author[2,3]{Namwoo Kang$^\dagger$}

\affil[1]{Department of Mechanical Engineering, KAIST}
\affil[2]{Cho Chun Shik Graduate School of Mobility, KAIST}
\affil[3]{Narnia Labs}

\begin{document}

\maketitle

\begingroup
\renewcommand{\thefootnote}{$^*$} 
\footnotetext{Equal contribution}
\renewcommand{\thefootnote}{$^\dagger$} 
\footnotetext{Corresponding author: nwkang@kaist.ac.kr}
\endgroup

\begin{abstract}
Generative Design (GD) has evolved as a transformative design approach, employing advanced algorithms and AI to create diverse and innovative solutions beyond traditional constraints. 
Despite its success, GD faces significant challenges regarding the manufacturability of complex designs, often necessitating extensive manual modifications due to limitations in standard manufacturing processes and the reliance on additive manufacturing, which is not ideal for mass production.
Our research introduces an innovative framework addressing these manufacturability concerns by integrating constraints pertinent to die casting and injection molding into GD, through the utilization of 2D depth images.
This method simplifies intricate 3D geometries into manufacturable profiles, removing unfeasible features such as non-manufacturable overhangs and allowing for the direct consideration of essential manufacturing aspects like thickness and rib design. 
Consequently, designs previously unsuitable for mass production are transformed into viable solutions.
We further enhance this approach by adopting an advanced 2D generative model, which offer a more efficient alternative to traditional 3D shape generation methods. 
Our results substantiate the efficacy of this framework, demonstrating the production of innovative, and, importantly, manufacturable designs.
This shift towards integrating practical manufacturing considerations into GD represents a pivotal advancement, transitioning from purely inspirational concepts to actionable, production-ready solutions. 
Our findings underscore usefulness and potential of GD for broader industry adoption, marking a significant step forward in aligning GD with the demands of manufacturing challenges.
\end{abstract}


\section{Introduction}

Generative Design (GD) represents a transformative approach in design methodology, leveraging algorithms, artificial intelligence (AI), and computational techniques to generate a vast array of design possibilities. 
This paradigm shift enables the exploration of numerous permutations far beyond traditional design methods, where outcomes are typically constrained by the designer's experience and the inherent limitations of manual processes. 
To date, GD has achieved remarkable success in producing diverse and innovative designs, optimizing for mechanical performance and aesthetic criteria \parencite{oh2019GD, yu2019deepTO, nie2021topologyGAN, yoo2021wheel, maze2023topodiff, giannone2023diff, regenwetter2022GDreview}.

However, a critical issue has emerged concerning the manufacturability of these complex designs. 
The intricacies and unconventional geometries frequently produced by GD pose significant challenges for traditional manufacturing processes such as die-casting or injection molding \parencite{alderighi2022moldreview}. 
Consequently, additive manufacturing (AM) techniques have become the go-to solution for fabricating GD-generated shapes \parencite{johnson20183dTOforAM, chen2023TOforAM}, despite AM's limitations for large-scale production due to cost and time constraints \parencite{leary2019DesignforAM, gibson2021DesignforAM}. 
This disconnect between design generation and mass-manufacturability necessitates substantial manual revisions by designers, diminishing GD's practicality for industrial applications. 
Thus, the current role of GD is predominantly inspirational, providing conceptual direction rather than actionable, production-ready solutions.

\begin{figure*}[t]
    \centering
    \includegraphics[width=\linewidth]{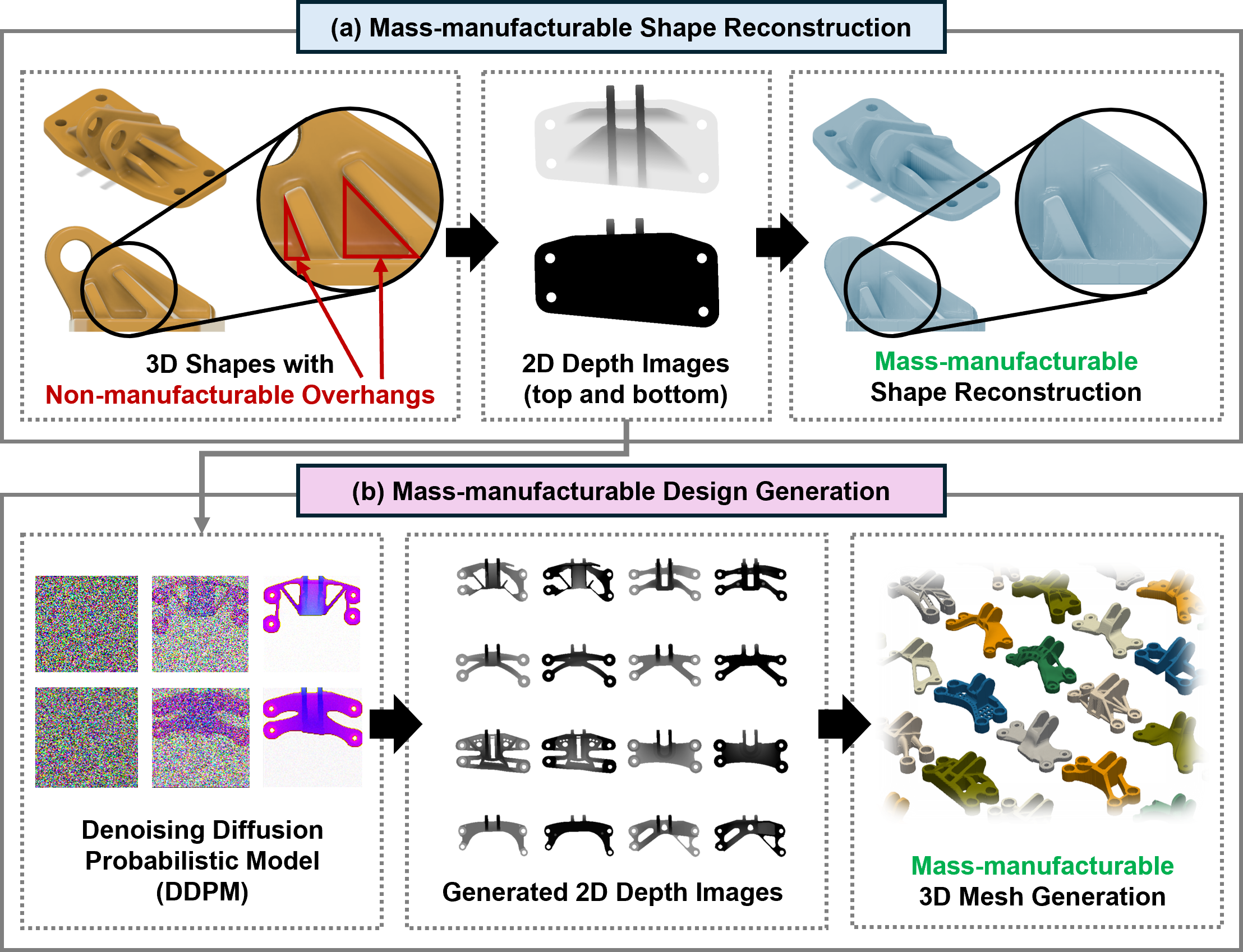}
    \caption{The framework of the proposed method, highlighting its capability to: (a) transform 3D shapes into designs suitable for mass production by using reconstructions from two 2D depth images, (b) while also leveraging powerful 2D generative models to create novel, diverse, and manufacturable 3D designs}
    \label{fig:framework}
\end{figure*}

Our paper introduces a simple but innovative framework in Fig.~\ref{fig:framework} that addresses the manufacturability concerns in GD by incorporating constraints relevant to die casting and injection molding, utilizing 2D depth images. 
These images simplify complex 3D geometries into manufacturable profiles by mapping the designs onto top and bottom plane projections. 
This process not only eliminates impractical features, such as overhangs, but also facilitates the consideration of critical factors for manufacturing like thickness and rib design directly through depth values. 
By transforming previously non-manufacturable designs into ones suitable for mass production, our method integrates advanced 2D generative models, including VAE, GAN, and diffusion probabilistic models, to efficiently generate innovative and diverse designs rather than relying on slow and heavy 3D models. 
Our method, which aligns GD with practical manufacturing needs, represents a significant shift from generating mere conceptual ideas to providing actionable, manufacturable solutions, thereby encouraging wider industry adoption. 
This combined focus on manufacturability, efficiency, and innovation through the use of 2D depth images marks a considerable advancement in making GD a practical tool for contemporary manufacturing challenges.

In summary, our contributions are as follows:
\begin{itemize}
\item Proposing a framework to incorporate manufacturability constraints, such as die casting and injection molding, into GD using 2D depth images
\item Simplifying complex 3D designs into manufacturable 2D profiles, addressing key manufacturing challenges such as overhangs
\item Employing advanced generative models on 2D depth images to enhance design diversity and efficiency
\item Demonstrating how our approach streamlines GD for practical manufacturing applications, promoting wider industry adoption and innovation
\end{itemize}

\section{Shape Reconstruction}

\begin{figure*}[tb]
    \centering
    \includegraphics[width=\linewidth]{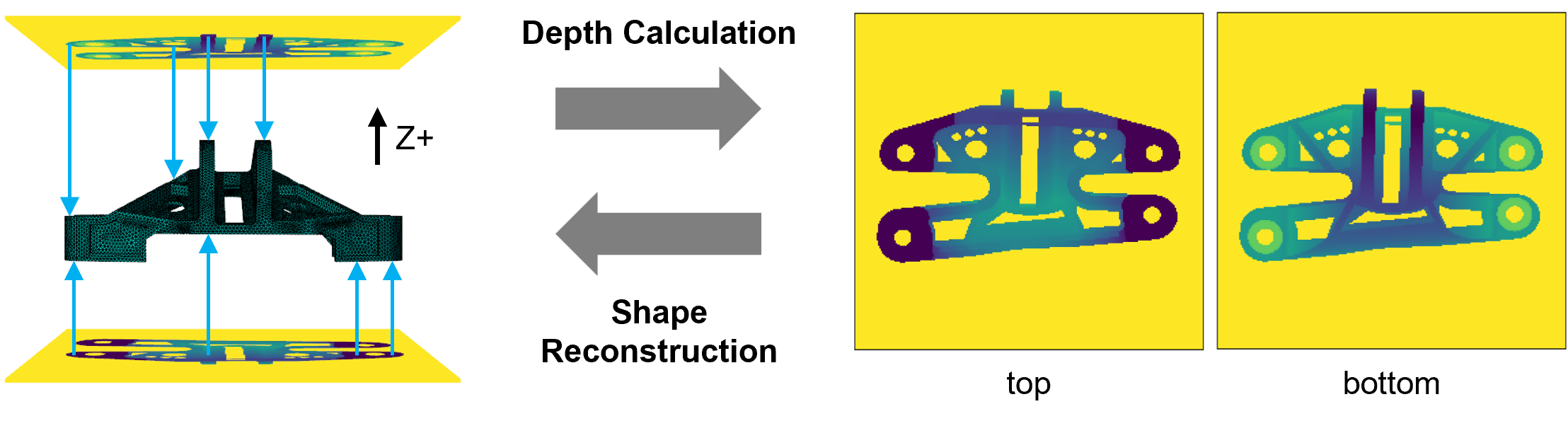}
    \caption{Calculating the distances between the top and bottom planes allows for the creation of 2D depth grids representing 3D shapes. This process facilitates the reconstruction of shapes, inherently removing any overhangs that cannot be manufactured.}
    \label{fig:depth-calc-and-shape-recon}
\end{figure*}

\subsection{Dataset}

\begin{figure}[htb]
    \centering
    \includegraphics[width=\linewidth]{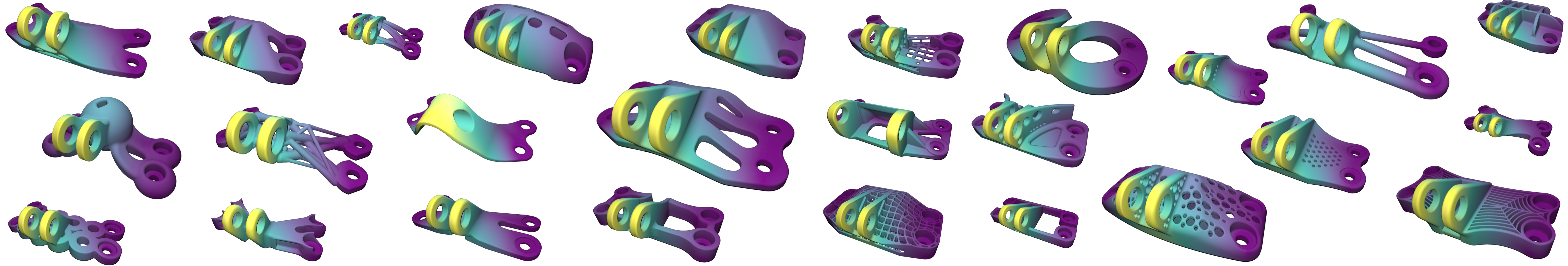}
    \caption{SimJEB engine bracket dataset \parencite{whalen2021simjeb}}
    \label{fig:SimJEB}
\end{figure}

In our study, we employed the SimJEB engine bracket dataset \parencite{whalen2021simjeb} shown in Fig.~\ref{fig:SimJEB} for the analysis. 
To ensure consistency and relevance to our manufacturing constraints, we filtered the dataset to include only those models that shared similar boundary conditions, which are the mounting holes in this dataset. 
This filtering process reduced the original dataset size from 381 to 264 models.

We stress that our methodology is adaptable to any 3D CAD or mesh-based datasets. 
Its flexibility in handling complex geometries and applying manufacturability constraints via 2D depth images makes it broadly applicable across various design and manufacturing domains.

\subsection{Depth Calculation}\label{ssec:depth-calc}

The z-axis was chosen as the primary direction, as illustrated in Fig~\ref{fig:depth-calc-and-shape-recon}. This selection was based on the observation that the cross-section of most shapes in the dataset was largest when perpendicular to the z-axis, thus enabling the straightforward application of our research findings. It is important to note, however, that the optimal axis may vary for different objects.

\begin{figure}[tb]
    \centering
    \includegraphics[width=\linewidth]{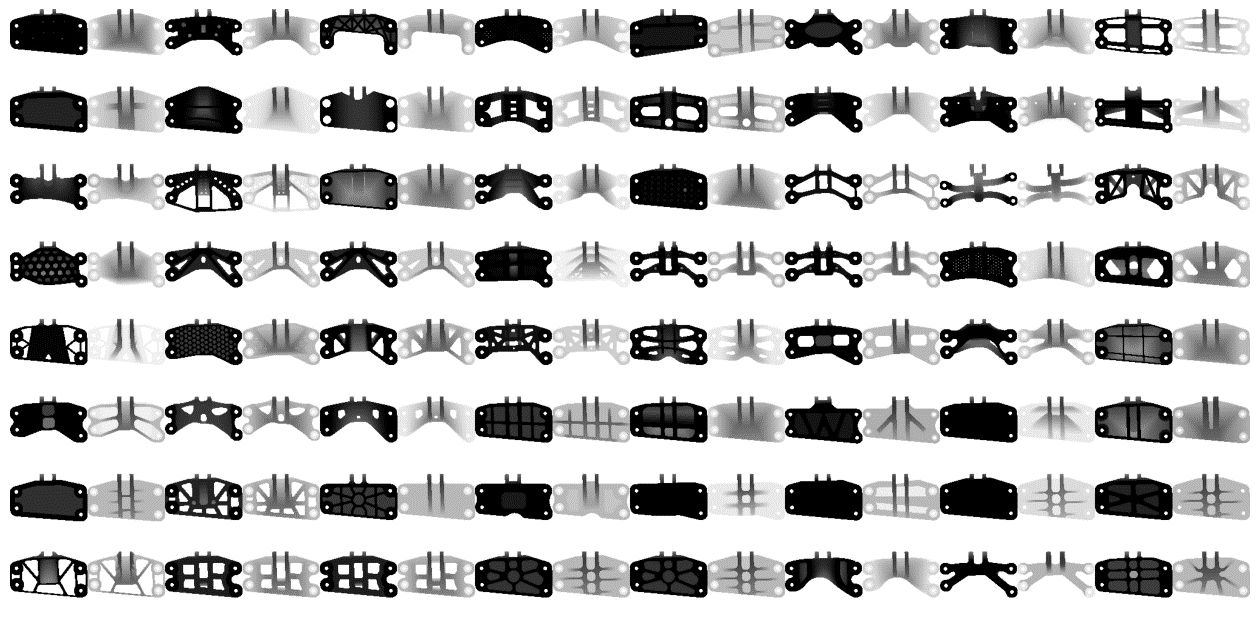}
    \caption{A subset of the calculated depth images}
    \label{fig:depth-images-demo}
\end{figure}

To perform the depth calculations necessary for our approach, we established 2D discrete grids with a resolution of 256, positioned in planes perpendicular to the z-axis. 
The depth at each point on these grids was determined by launching rays in the z-axis direction, from both the top and the bottom surfaces of the engine bracket shapes.
Specifically, we calculated depth values at z = 0.4 for the top plane and z = -0.4 for the bottom plane, within a space defined by an implicit representation. 
This method of using implicit representations for depth calculation ensures greater accuracy in the resulting depth values, presented in float format. 
It offers a significant advantage over voxelization techniques, which tend to yield less precise values unless extremely high resolutions are employed. 
A subset of the obtained depth images are shown in Fig.~\ref{fig:depth-images-demo}.

The depth images obtained from scanning the shapes can be considered as an inverse of the mold design typically used in die casting or injection molding processes. 
This perspective allows us to directly relate the geometric features captured in the depth images to practical mold designs, enhancing the applicability of our results to the manufacturing field.
We refer to a comprehensive review of the latest computational mold design techniques by \cite{alderighi2022moldreview} for more information.

\subsection{Manufacturable Shape Reconstruction}

The generative model employed in our study produces depth images, where each pixel is assigned a specific depth value, representing the in-space heights from the predefined top and bottom planes. 
This depth information serves as a critical component in reconstructing manufacturable shapes. 
By projecting these depth values from the top and bottom planes, we can reconstruct a three-dimensional shape that adheres to the manufacturability constraints integral to the die casting or injection molding processes as shown in Fig.~\ref{fig:depth-calc-and-shape-recon}.

\begin{figure}[tb]
    \centering
    \includegraphics[width=\linewidth]{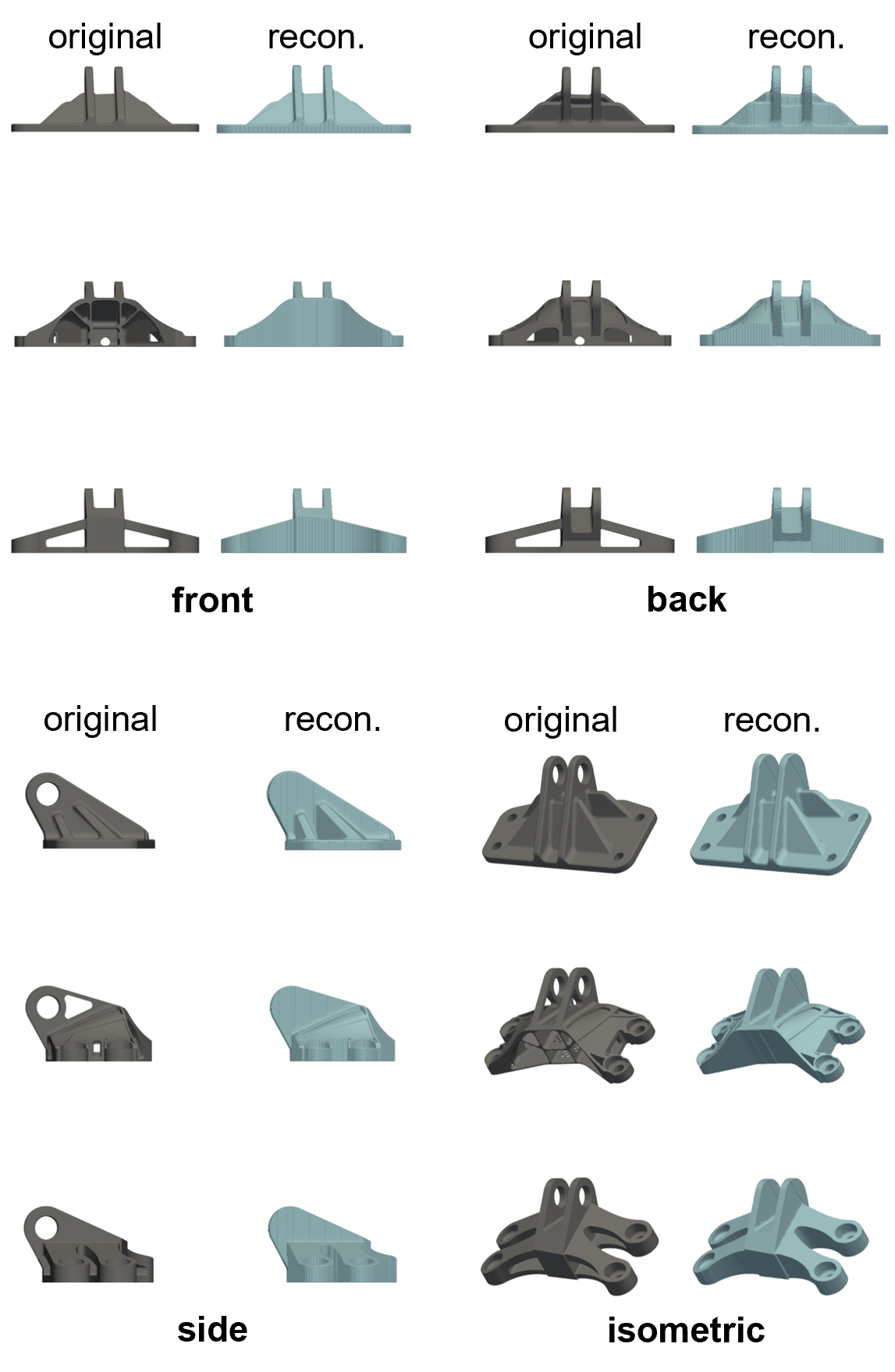}
    \caption{Illustration of the original and the reconstructed (recon.) shapes. The top and the bottom are identical.}
    \label{fig:results-recon}
\end{figure}

An innovative aspect of our approach is that the overall shape of the designed object can be effectively defined using just two depth images oriented in the casting direction. 
This method naturally eliminates any non-manufacturable overhangs, which are features problematic for traditional mass production processes. 
By doing so, it ensures that all elements of the reconstructed shape are manufacturable.
The reconstruction results are shown in Fig.~\ref{fig:results-recon}.

For intricate features such as holes that may occur in the y-axis of the bracket, we incorporate a post-processing step. 
Boolean operations are applied to the resultant mesh to introduce these features accurately. This allows for the inclusion of necessary functional aspects, such as mounting holes for the engine brackets, without compromising the manufacturability imposed by the constraints of traditional casting techniques. 
By integrating this post-processing step, our methodology ensures that the final reconstructed shapes are not only optimized for manufacturing but also meet the required design specifications.


\section{Design Generation}
Our approach utilizes Denoising Diffusion Probabilistic Models (DDPMs) \parencite{ho2020ddpm} to create 2D depth images, facilitating the generation of manufacturable designs while retaining a high degree of diversity and complexity. 
This section outlines the methodology employed for design generation within our framework.

\subsection{Denoising Diffusion Probabilistic Models}
DDPMs are a class of generative models that learn to gradually denoise a signal, starting from a random distribution and moving towards the distribution of the target data. 
In our case, the target data consists of 2D depth images representing the top, bottom, and edge profiles of designs subject to manufacturing constraints.

The DDPM process can be mathematically described in two phases: the forward process and the reverse process.

\paragraph{Forward Process}
The forward process, also known as the diffusion process, involves adding Gaussian noise to the data over a series of steps, transforming the original data \( x_0 \) (our depth images) into a completely noisy state \( x_T \). This can be represented as:

\begin{equation}\label{eq:diffusion-forward}
    x_t = \sqrt{\alpha_t} x_0 + \sqrt{1 - \alpha_t} \epsilon_t
\end{equation}
where \( t \) ranges from 1 to \( T \), \( \alpha_t \) are fixed variance schedules, and \( \epsilon_t \) is sampled from a standard normal distribution. This process generates a sequence of increasingly noisy images.

\paragraph{Reverse Process}
The reverse process, or denoising process, aims to learn the distribution of the original data by reversing the diffusion process. It is modeled by a neural network that parameterizes the conditional distribution \( p(x_{t-1} | x_t) \). The objective is to gradually denoise the images, starting from noise \( x_T \) and moving towards the data distribution to generate a clean image \( x_0 \):

\begin{equation}\label{eq:diffusion-reverse}
x_{t-1} = \frac{1}{\sqrt{\alpha_t}} \left(x_t - \frac{1-\alpha_t}{\sqrt{1 - \alpha_t}} \epsilon_\theta(x_t, t)\right) + \sigma_t z
\end{equation}
where \( \epsilon_{\theta}(x_t, t) \) is the output of the neural network, providing an estimate of the noise, and \( z \) is a noise term that is gradually phased out as \( t \) decreases.

\subsection{Adaptation to Depth Images}

Our approach adapts DDPMs to the domain of depth images by considering them in floating-point format instead of the traditional integer format for bitmap images. 
This allows us to capture more detailed gradients and subtleties in the depth profiles, which is crucial for accurate representation of manufacturable designs.

\subsection{Edge Detection}

\begin{figure}[h!]
    \centering
    \includegraphics[width=\linewidth]{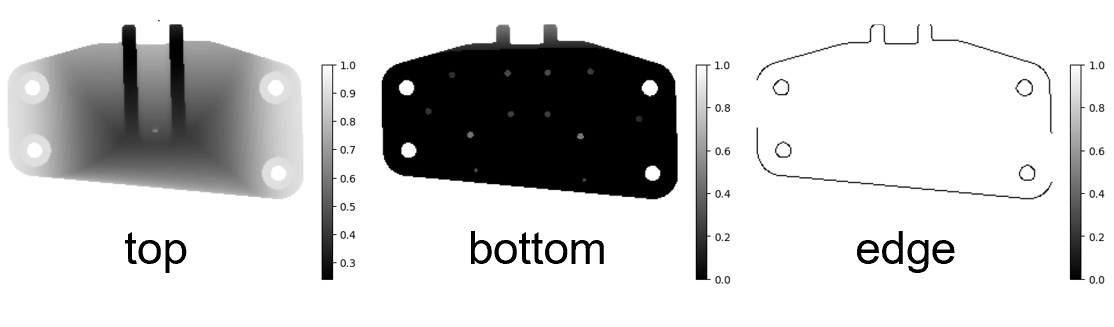}
    \caption{The top, bottom and edge channels used to train the design generation model}
    \label{fig:3ch-demo}
\end{figure}

In our framework, edges within the depth images are highlighted using Canny edge detector \parencite{canny1986edge} to emphasize critical geometric features. 
We incorporate these edge features directly into the diffusion process along with the top and bottom depth images, allowing the model to better understand and generate the important boundaries and interfaces of the designs, while conserving the continuity between the top and the bottom shapes.
This three-channel representation consists of top, bottom, and edge details shown in Fig.~\ref{fig:3ch-demo} enables high quality generation.
We stress that the edge channel is only used to aid the training process, not for the shape reconstruction.

\subsection{Training and Generation}
The diffusion model is trained on a dataset of the depth images obtained in Sec.~\ref{ssec:depth-calc}. 
The model learns to generate new designs that inherently respect these constraints, thus ensuring manufacturability without compromising design diversity.
The training of the DDPM involves optimizing the neural network to accurately estimate the noise in the diffusion process, thereby improving the quality and accuracy of the generated designs. 

\begin{figure}[h!]
    \centering
    \includegraphics[width=\linewidth]{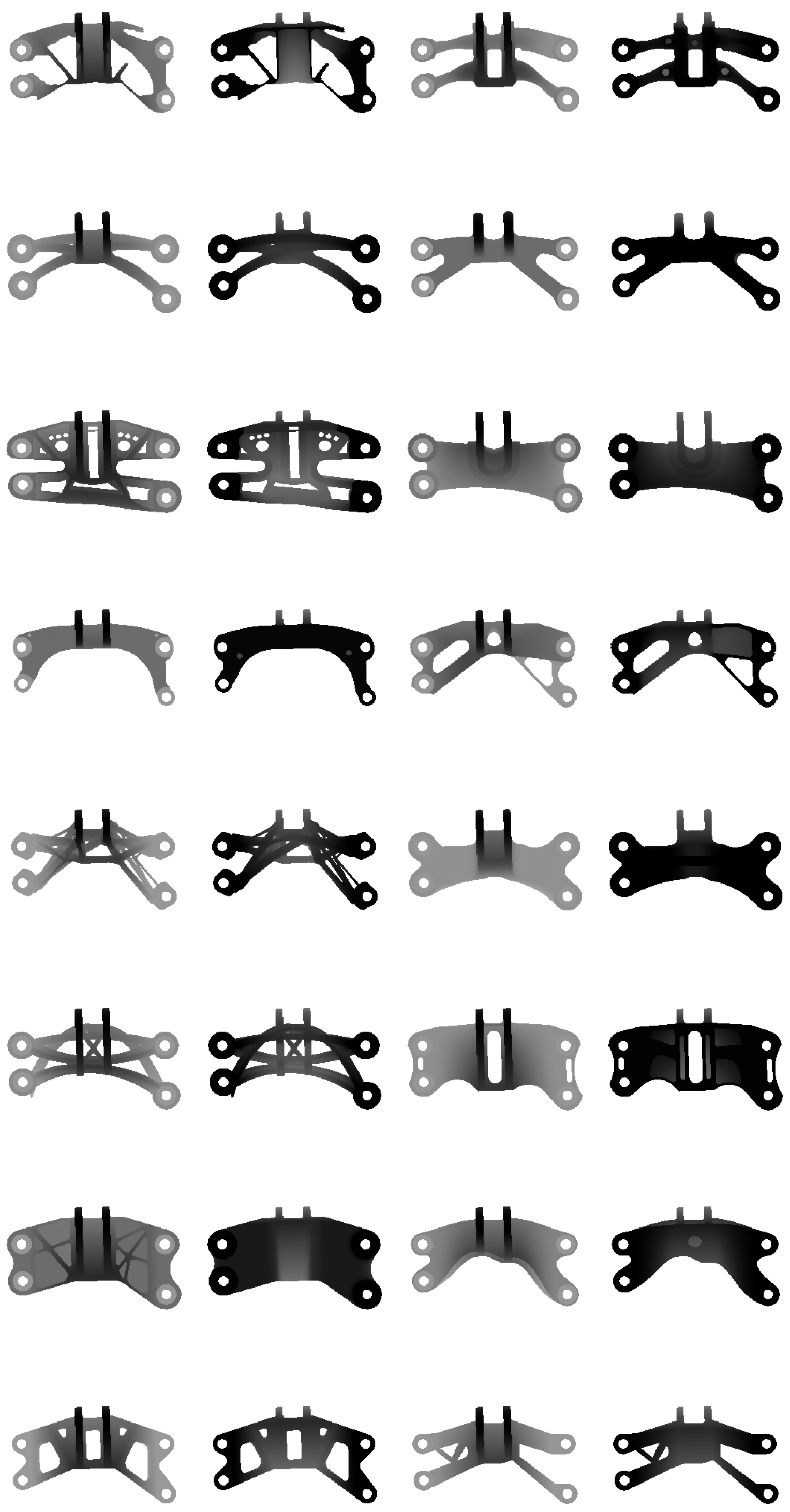}
    \caption{Depth images generated by DDPM}
    \label{fig:results-diffusion}
\end{figure}

\begin{figure*}[h!]
    \centering
    \includegraphics[width=\linewidth]{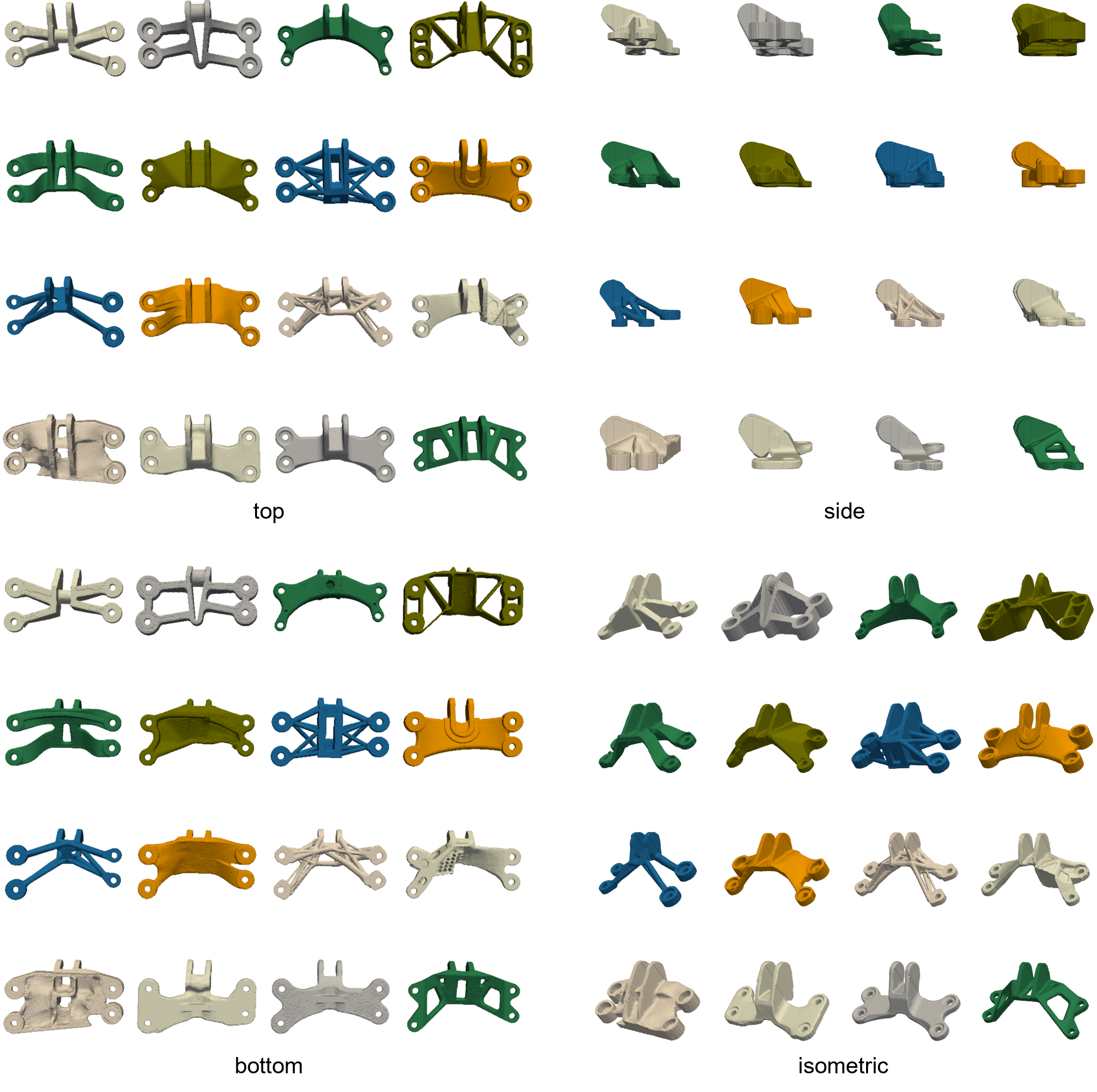}
    \caption{3D mesh generation using the depth images generated in Fig.~\ref{fig:results-diffusion}}
    \label{fig:results-diffusion-recon}
\end{figure*}

After the training, the model can generate new depth images by sampling from the noise distribution and iteratively applying the learned denoising steps.
The generated depth images are shown in Fig.~\ref{fig:results-diffusion}. 
These generated images are then used to generate 3D models as shown in Fig.~\ref{fig:results-diffusion-recon}.

\section{Conclusion}\label{sec:conclusion}
\subsection{Summary}
Our study marks a significant advancement in melding GD with practical mass production techniques, particularly die casting and injection molding. 
Through the innovative integration of manufacturability constraints via 2D depth images, we bridge a crucial gap between imaginative GD concepts and their viability for real-world mass production. 
Our approach streamlines the complex translation from intricate 3D geometries to manufacturable 2D profiles while leveraging the efficiency and diversity of advanced 2D generative models. 
The significant outcomes of our research, which highlight reduced computational costs, shortened timelines from design to production, and the generation of more diverse, innovative designs, underscore the profound impact of incorporating practical manufacturing considerations into GD, enhancing its utility with broader industry adoption. 
We argue that this methodology represents a crucial step forward, transforming GD from a mere conceptual tool into a practical, efficient, and innovative solution for the manufacturing challenges of the future.

\subsection{Limitations and Future Work}

The design outcomes in our study have been influenced by the limited diversity of our dataset, which includes less than 300 images. This limitation has resulted in generated designs that resemble combinations of multiple existing brackets. To enhance the uniqueness and applicability of generated designs, future research should focus on expanding the dataset to include a broader range of images. This enhancement would likely yield more diverse and innovative design outputs.

\subsection*{Acknowledgements}
This work was supported by the National Research Foundation of Korea grant (2018R1A5A7025409) and the Ministry of Science and ICT of Korea grant (No.2022-0-00969, No.2022-0-00986).

\printbibliography
\end{document}